\documentclass[10pt,twocolumn,letterpaper]{article}

\usepackage[pagenumbers]{cvpr} % To force page numbers, e.g. for an arXiv version

\usepackage[dvipsnames,table]{xcolor}
\usepackage[pagebackref,breaklinks,colorlinks,allcolors=cvprblue]{hyperref}
\definecolor{cvprblue}{rgb}{0.21,0.49,0.74}

\definecolor{darkgreen}{RGB}{0,127,0}
\definecolor{darkred}{RGB}{200,0,0}
\definecolor{pink}{RGB}{255, 50, 203} 
\definecolor{blue}{rgb}{0, 0, 1}

\newcommand{\pink}[1]{{\color{pink}#1}}

\usepackage{times}
\usepackage{amsmath}
\usepackage{amssymb}
\usepackage[accsupp]{axessibility}  % CVPR Request

\usepackage{graphicx}  % epsfig 대신 graphicx만 사용
\usepackage{verbatim}

\usepackage{booktabs}  % 중복 제거
\usepackage{multirow}
\usepackage{makecell}
\usepackage{bigstrut}
\usepackage{hhline}
\usepackage{pifont}  % for \ding{55}
\usepackage[font=footnotesize,skip=10pt]{caption} 
\usepackage{xspace}

\newcommand{\Tref}[1]{Table~\textcolor{blue}{\ref{#1}}}

\newcommand{\Fref}[1]{Fig.~\textcolor{blue}{\ref{#1}}}
\newcommand{\Sref}[1]{Sec.~\textcolor{blue}{\ref{#1}}}

\def\greencheckmark{\textcolor{darkgreen}{\checkmark}}
\def\redxmark{\textcolor{darkred}{\ding{55}}}

\newcommand{\thickline}[0]{\Xhline{3pt}}

\newcommand{\link}[1]{\href{#1}{#1}}

\def\paperID{} % *** Enter the Paper ID here
\def\confName{CVPR}
\def\confYear{2025}

\title{Any6D: Model-free 6D Pose Estimation of Novel Objects}

\author{Taeyeop Lee$^{1}$ \quad Bowen Wen$^{2}$ \quad Minjun Kang$^{1}$ \quad Gyuree Kang$^{1}$ \quad In So Kweon$^{1}$ \quad Kuk-Jin Yoon$^{1}$\\ \\
$^{1}$KAIST~~~~~~$^{2}$NVIDIA~~
}

\begin{document}
\maketitle
\begin{abstract}
We introduce Any6D, a model-free framework for 6D object pose estimation that requires only a single RGB-D anchor image to estimate both the 6D pose and size of unknown objects in novel scenes. Unlike existing methods that rely on textured 3D models or multiple viewpoints, Any6D leverages a joint object alignment process to enhance 2D-3D alignment and metric scale estimation for improved pose accuracy. Our approach integrates a render-and-compare strategy to generate and refine pose hypotheses, enabling robust performance in scenarios with occlusions, non-overlapping views, diverse lighting conditions, and large cross-environment variations. We evaluate our method on five challenging datasets: REAL275, Toyota-Light, HO3D, YCBINEOAT, and LM-O, demonstrating its effectiveness in significantly outperforming state-of-the-art methods for novel object pose estimation. Project page: \link{https://taeyeop.com/any6d}
\end{abstract}    
\section{Introduction}
\label{sec:intro}

% Pose Estimation
Object 6D pose estimation is a crucial problem in computer vision and robotics, focusing on determining the rigid 6D transformation, comprising both 3D orientation and 3D translation, between reference and camera coordinates. This task has numerous practical applications, such as robotic manipulation~\cite{wen2022you, zeng2017multi, wong2017segicp, wang2019densefusion, du2021vision, blukis2023oneshot} and augmented reality~\cite{runz2018maskfusion, marchand2015pose, marder2016project}. In recent years, significant progress has been made in this field~\cite{hodan2024bop, liu2024deep, du2021vision, banerjee2024introducing}, and research is ongoing.

% Object Pose Estimation
Research in 6D object pose estimation can be broadly categorized into three approaches: instance-level~\cite{kehl2017ssd, tremblay2018corl:dope, peng2019pvnet,wen2020robust}, category-level~\cite{wang2019normalized, Tian2020prior, lin2022icra:centerpose, lee2021category, lee2022uda, lee2023tta}, and category-agnostic methods~\cite{he2022fs6d, wen2024foundationpose, nguyen2024gigapose, ornek2024foundpose}. 
Instance-level methods provide high precision but come with significant limitations. They rely on exact RGB-textured CAD models for estimating object poses, which restricts their effectiveness to only those objects seen during training. A major drawback is their inability to handle new objects without additional fine-tuning. Category-level methods partially mitigate these limitations by using category-specific prior knowledge, but they are still restricted to predefined object categories. Moreover, they face considerable challenges in acquiring comprehensive training datasets due to the complexity of aligning canonical poses. In contrast, category-agnostic methods aim to generalize across different objects without being limited to specific instances or categories, offering a more flexible approach to 6D pose estimation.

\begin{figure}[t]
\begin{center}
\includegraphics[width=1.0\linewidth]{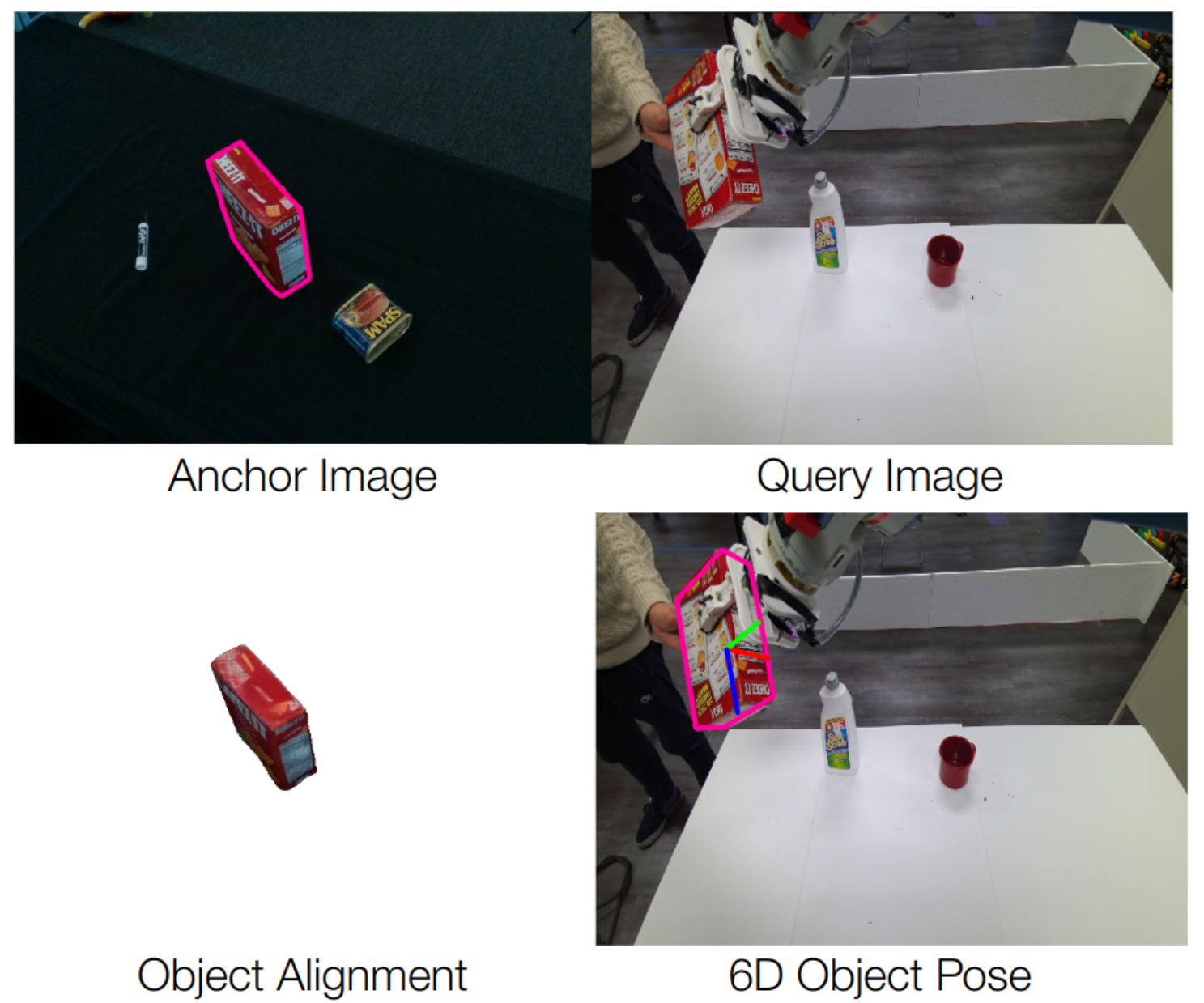}
\vspace{-0.2in}
\caption{Our method accurately estimates 6D object pose for novel objects on drastically different scenes and viewpoints using only a single RGB-D anchor image. We achieve robust pose estimation without requiring precise CAD models or posed multi-view reference images.}
\label{fig:teaser}
\end{center}
\vspace{-0.3in}
\end{figure}

% Novel Object Pose Estimation
Recent research has shifted toward category-agnostic approaches~\cite{nguyen2024gigapose, pinar2023foundpose, lin2024sam, shugurov2022osop, wen2024foundationpose, labbe2022megapose, nguyen2024gigapose,wen2023bundlesdf} to address the limitations of both category-level and instance-level pose estimation. These efforts can be broadly divided into two directions: model-based methods~\cite{nguyen2024gigapose, lin2024sam, pinar2023foundpose}, which require textured RGB 3D CAD models at test time, and model-free methods~\cite{sun2021loftr, he2022fs6d, liu2022gen6d, he2022oneposepp}, which utilize multiview reference images or video sequences of the target object during inference. Although both approaches show promising results, they still face significant practical limitations when dealing with unseen objects that are not physically accessible. In robotic manipulation scenarios, for example, these methods face difficulties when a robot encounters unexpected objects in a new environment without available 3D models or multiview images. This dependency on prior object data significantly limits their effectiveness in real-world applications.

% Relative Pose Estimation from 
To address these challenges, Oryon~\cite{corsetti2024open} introduces a novel model-free approach for estimating the 6D pose of objects from a single reference RGBD image. Unlike earlier methods that rely heavily on detailed object data at test time, Oryon uses language guidance to perform pose estimation with just a single RGBD reference. This method demonstrates impressive adaptability, functioning effectively even when reference and target images come from entirely different environments. Oryon establishes correspondences between images using language cues and estimates the pose by aligning visible point clouds from the reference. However, its performance degrades when occlusions or minimal overlap object regions between the reference and target objects, limiting the number of matching points. Consequently, Oryon struggles in object manipulation scenarios involving human or robotic arms, particularly when targets are occluded, appear in non-overlapping views, or lack sufficient texture~\cite{wen2020se, hampali2020honnotate}, as shown in our experiments.

% Our work
To overcome these limitations, we present Any6DPose, a novel model-free approach for object pose estimation using a single anchor RGBD image. Inspired by recent advances in image-to-3D models~\cite{hong2023lrm, liu2024one, xu2024instantmesh, xiang2024structured, long2024wonder3d, wen2024ouroboros3d}, our method estimates metric scale object shape for object pose estimation. Although existing 3D generation methods achieved promising photorealistic consistency between the input image and the generated 3D shape in normalized space, they often neglect critical aspects of 2D-3D alignment, especially the metric scale, which is crucial for accurate pose estimation and the subsequent downstream tasks. Therefore, we introduce a simple yet effective object alignment, jointly improving object size and pose estimation by alignment in 2D and 3D space. 
We generate multiple pose hypotheses and use a render and compare strategy to select the optimal one, building on previous work~\cite{wen2024foundationpose}. This enables our method to effectively handle disjoint views between query and anchor objects, as well as 
occlusions, diverse light conditions, and large cross-environment variations. We validate our approach across diverse scenarios on five public datasets: HO3D, YCBINEOAT, REAL275, Toyota-Light, and LM-O. Our Any6DPose significantly outperforms state-of-the-art methods in novel object pose estimation.

The main contributions of our work are as follows:
\begin{itemize}
 \setlength\itemsep{-0.00em}
    \item We introduce Any6DPose, a novel framework that enables 6D pose and size estimation of novel objects in different scenes from only a single reference image.
    
    \item We propose a straightforward yet effective object alignment technique that addresses the challenges of existing 3D generation models, specifically improving 2D-3D alignment and size estimation for accurate pose estimation.

    \item We validate our approach through extensive experiments, demonstrating superior performance compared to state-of-the-art methods across five benchmark datasets.
    
\end{itemize}

\section{Related Works}
\label{sec:related}

\begin{figure*}[t]
\begin{center}
\vspace{-0.05in}
\includegraphics[width=1.0\linewidth]{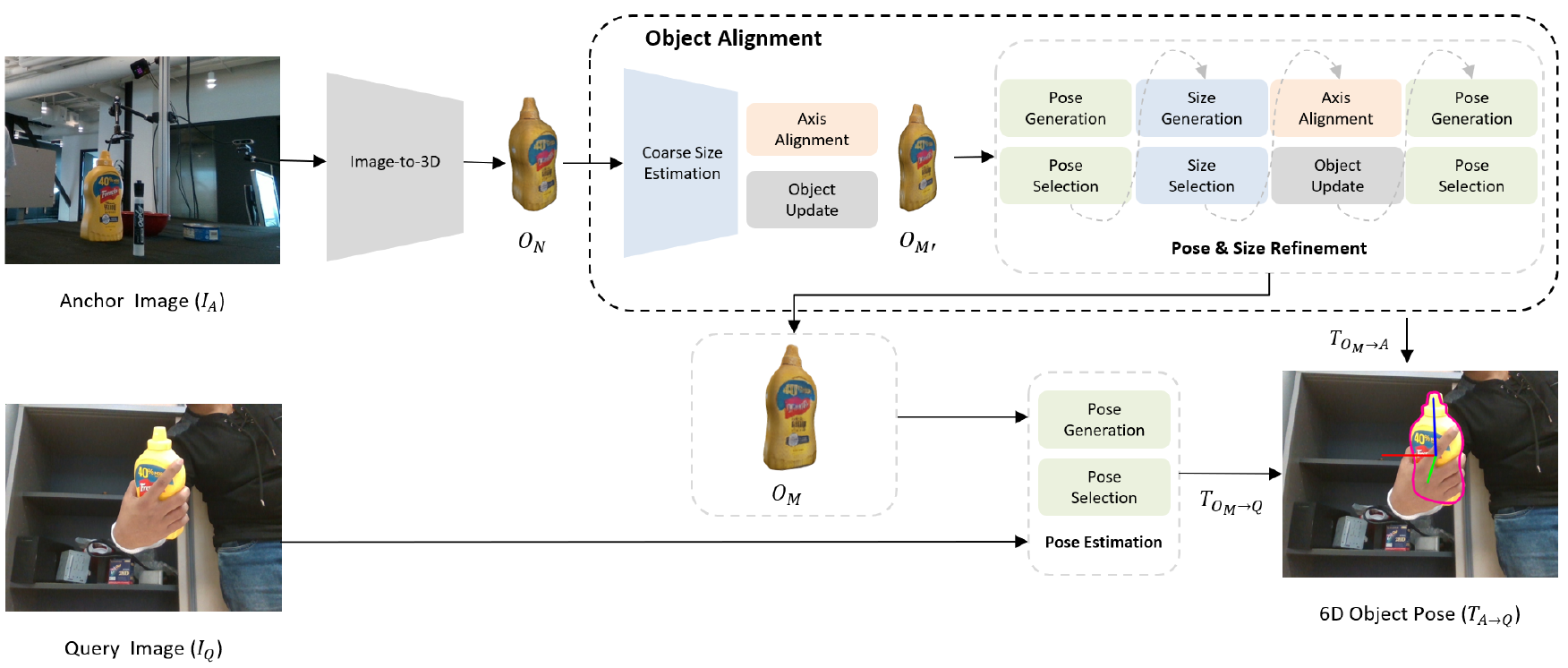}
\vspace{-0.2in}
\caption{Overview of the Any6D framework for model-free object pose estimation.
First, we reconstruct normalized object shape $O_N$ from the image-to-3D model. Then, we estimate accurate object pose and size from anchor image $I_A$ using the proposed object alignment  (\Sref{subsec:object_alignemt}). Next, we use the query image $I_Q$ to estimate the pose with the reconstructed metric-scale object shape $O_M$ (\Sref{subsec:pose_estimation}).
}
\label{fig:method}
\end{center}
\vspace{-0.25in}
\end{figure*}

\subsection{CAD Model-based Object Pose Estimation}
Instance-level pose estimation methods~\cite{wang2019densefusion, he2020pvn3d, he2021ffb6d, park2019pix2pose, labbe2020cosypose} rely on textured CAD models of specific objects, with training and testing conducted on the same instances.
Category-level methods~\cite{wang2019normalized, chen2020cass, lin2024instance, chen2024secondpose, zhang2024omni6dpose} generalize to novel instances within known categories but require expensive category annotations and remain constrained to predefined object classes.
Category-agnostic methods~\cite{lin2024sam, caraffa2024freeze, chen2024zeropose} address these limitations by estimating the poses of arbitrary novel objects without category constraints. However, most still utilize ground-truth CAD models during inference, restricting their practical application in real-world scenarios where such models are unavailable. Some recent works~\cite{ponimatkin2025d, liu2023openshape} try to retrieve CAD from existing databases.

\subsection{Model-Free Object Pose Estimation}
To overcome the limitations of CAD dependencies, model-free approaches have been developed that eliminate the need for explicit textured models by relying instead on a set of reference images of the target object. Gen6D~\cite{liu2022gen6d}, OnePose~\cite{sun2022onepose}, and OnePose++\cite{he2022oneposepp} generate 3D point clouds from videos or multiple views using structure-from-motion (SfM)~\cite{schonberger2016structure} and utilize these point clouds for pose estimation through 2D-3D matching~\cite{kneip2011novel, lepetit2009epnp}. FS6D~\cite{he2022fs6d} and FoundationPose~\cite{wen2024foundationpose} extend this approach to RGB-D images, achieving promising results. However, these methods still depend on multiview images, with most requiring camera poses to merge these views, which is often impractical in real-world scenarios.

Motivated by these challenges, recent work has focused on reducing these dependencies. For example, NOPE~\cite{nguyen2024nope} estimates relative orientation using only a single anchor image. LoFTR~\cite{sun2021loftr} also contributes by using a transformer-based approach for feature matching, effectively enhancing the robustness of pose estimation from a single image. Similarly, Oryon~\cite{corsetti2024open} reduces the requirement to a single image by incorporating language guidance.  However, these partial matching methods~\cite{sun2021loftr, corsetti2024open, poiesi2022learning, detone2018superpoint, liu2024unopose, liu2025novel} struggle in scenarios with non-overlap regions or significant occlusions in the target image. GigaPose~\cite{nguyen2024gigapose}, closely related to our approach, utilizes image-to-3D (Wonder3D~\cite{long2024wonder3d}) to estimate object pose but still requires an initial novel object pose to determine object size. 
Concurrently, HIPPo~\cite{liu2025hippo}, OmniManip~\cite{pan2025omnimanip}, and SceneComplete~\cite{agarwal2024scenecomplete} explore model-free object pose estimation from an image using image-to-3D methods.

\section{Method}
Given an RGB-D anchor image ($I_A$) and an RGB-D query image ($I_Q$), our task is to estimate the relative pose between them. The query images may capture the same object from drastically different viewpoints and scenes from the single anchor image. We formulate the problem as a relative pose estimation task~\cite{sun2021loftr, poiesi2022learning, bai2021pointdsc, corsetti2024open}. Our method aims to estimate the relative 6D pose $\mathbf{T}_{A \rightarrow Q} \in \mathrm{SE}(3)$ between $I_A$ and $I_Q$, where $\mathbf{T}_{A \rightarrow Q}$ is defined as the rigid transformation $[R \mid t]$, consisting of a rotation $R \in \mathrm{SO}(3)$ and a translation $t \in \mathbb{R}^3$.

Previous approaches have attempted matching using either visible RGB images~\cite{sun2021loftr, sarlin20superglue} or point clouds~\cite{bai2021pointdsc, poiesi2022learning}. While effective with significant overlap, these methods struggle in scenarios with occlusions or large viewpoint variations, as partial-to-partial matching lacks sufficient shared features. To address these challenges, our method adopts a full-to-partial matching strategy by reconstructing a complete 3D object shape, ensuring robust alignment even under little overlap. Accurate pose estimation is achieved through a render-and-compare pipeline~\cite{wen2024foundationpose}, effectively handling occluded or partially visible objects. The benefits of our approach are confirmed through extensive experiments (\Sref{subsec:sota_comparision}), demonstrating superior performance in low-overlap scenarios.

We introduce Any6D, a framework for estimating the relative pose $\mathbf{T}_{A \rightarrow Q}$ between an anchor image $I_A$ and a query image $I_Q$. Our framework comprises two components: first, we reconstruct the normalized object shape $O_N$ from the anchor image using an image-to-3D model, without considering real-world scale or pose, and then estimate the metric-scale object shape $O_M$ by determining both the actual object size $s \in \mathbb{R}^3$ and the pose $T_{O_M \rightarrow A}$, aligning it properly in 2D and 3D space (\Sref{subsec:object_alignemt}). Next, we use the reconstructed metric-scale object shape and the query image to estimate the pose, deriving the relative transformation $\mathbf{T}_{A \rightarrow Q}$ by combining $T_{O_M \rightarrow A}$ and $T_{O_M \rightarrow Q}$ (\Sref{subsec:pose_estimation}). The complete workflow of our Any6D framework is illustrated in \Fref{fig:method}.

\subsection{Coarse Object Alignment}\label{subsec:object_alignemt}
To our knowledge, there is no reliable existing solution for single-view metric-scale reconstruction from RGB-D that can handle diverse objects effectively. Given the recent advancements in RGB-based single-view reconstruction~\cite{xu2024instantmesh, hong2023lrm, liu2023zero}, we thus resort to InstantMesh~\cite{xu2024instantmesh} which has shown promising results across various objects. One key limitation, however, is the 3D object reconstruction only yields a shape with a normalized scale ${O}_{N}$, in the range [-1, 1] for each XYZ axis, meaning the resulting meshes are not properly scaled or positioned relative to the actual scene. This limitation prevents us from obtaining accurate pose alignment further, thus motivating our object alignment step, where we first estimate a coarse size of the object shape ${O}_{M}$ and then refine this size by jointly solving accurate pose $T_{O_M \rightarrow A}$. Our approach involves estimating and aligning object shapes in both 3D and 2D spaces between $I_A$ and $O_N$, including $\mathbf{T}_a \in \mathrm{SE}(3)$, and size $s\in \mathbb{R}^3$.

\begin{figure}[t]
\begin{center}
\includegraphics[width=1.0\linewidth]{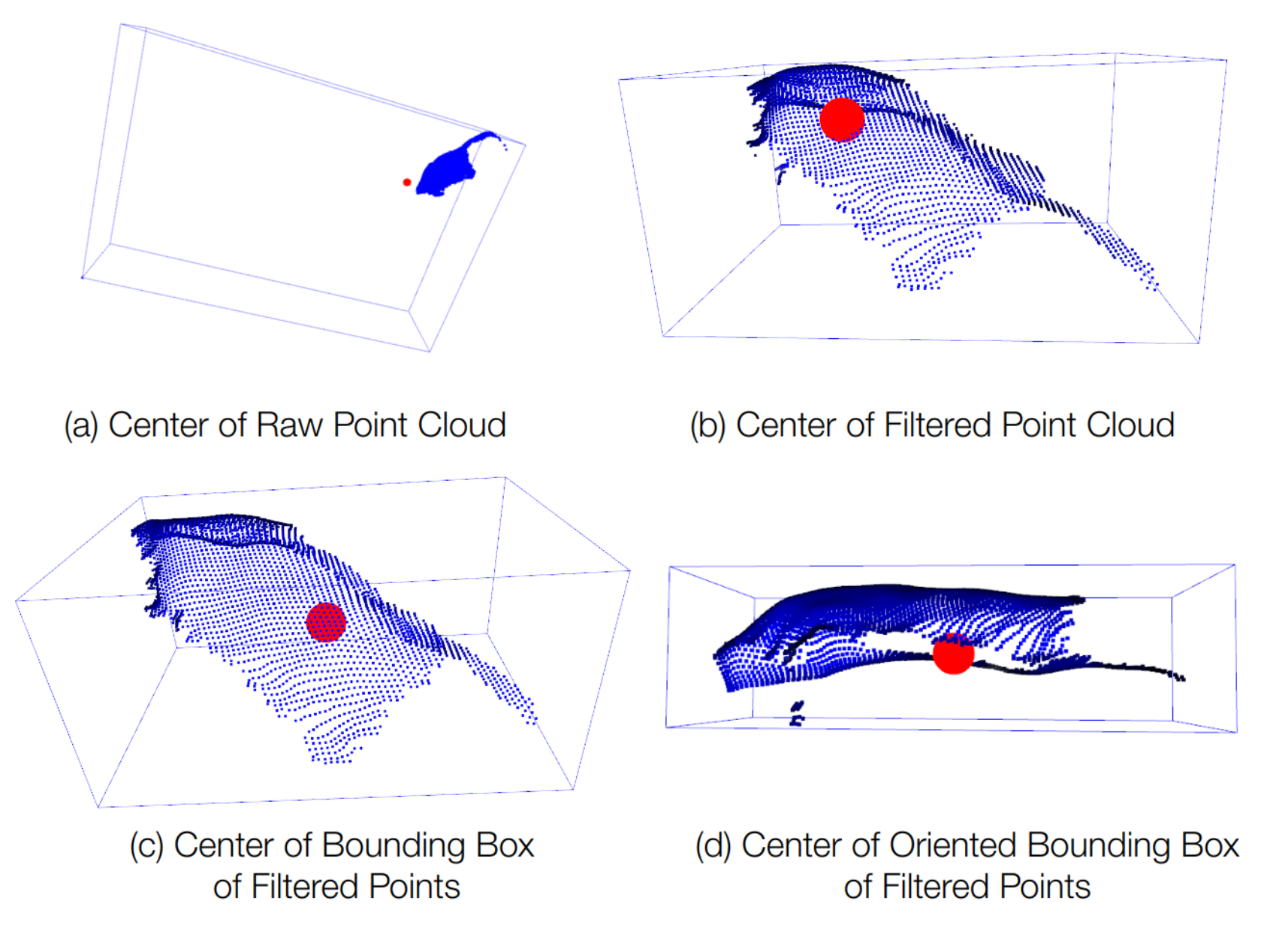}
\vspace{-0.2in}
\caption{Visualization of each point clouds and center of mustard object.}
\label{fig:points}
\end{center}
\vspace{-0.3in}
\end{figure}

Specifically, we estimate the object size $s$ in a coarse-to-fine manner using $I_A$. We first initialize the coarse object size by comparing point clouds between $I_A$ and ${O}_{N}$ from their respective object centers. While the mean of points is a straightforward approach for center estimation, it becomes unreliable due to partial viewpoint and noisy outlier points in the anchor image, as shown in~\Fref{fig:points}-(a, b). Using a simple axis-aligned bounding box also leads to inaccurate center estimation due to the partial observability, as shown in~\Fref{fig:points}-(c).  Therefore, we propose using an oriented bounding box to determine the object center, as shown in~\Fref{fig:points}-(d), which yields a more reliable coarse center estimate for $I_A$.
For the axis alignment, we align the oriented bounding box with the XYZ axis.
We then sample various rotation angles and calculate the IoU between the rotated bounding boxes of $I_A$ and ${O}_{N}$ across different angles. The combination of rotation and scale that leads to the highest IoU is used to transform ${O}_{N}$ into a coarsely aligned object shape, updating it to an initial object shape ${O}_{M'}$, which is subsequently used for accurate pose and size estimation.

\subsection{Fine Object Alignment}\label{subsec:pose_estimation}
Our framework aims to determine the relative pose $\mathbf{T}_{A \rightarrow Q}$ between an anchor image $I_A$ and a query image $I_Q$. This process involves multiple steps, starting with the reconstruction of a coarse object shape ${O}_{M'}$ and subsequently refining it to obtain the final metric-scale object shape ${O}_{M}$. Given the coarsely scaled initial shape ${O}_{M'}$, we jointly refine both the pose and the object size through our object and size refinement.

To refine the object shape and pose, we draw inspiration from FoundationPose~\cite{wen2024foundationpose} for its effective pose generation, refinement, and selection capabilities. However, it either requires ground-truth metric-scale object CAD for its model-based setup or multiple posed reference images in its model-free setup. This prevents its direct application in our considered setup, where only a single RGBD reference image is provided. We thus develop a joint module that injects the size estimation task into its pose refinement process. This allows us to estimate both the metric-scale size and pose reliably simultaneously.

Our refinement pipeline involves three primary modules: pose estimation, size estimation, and axis alignment—all of which work together in a unified process by alternating between the task of size refinement and pose refinement. We begin by estimating an initial pose using ${O}_{M'}$ and simultaneously refining the object size. In FoundationPose~\cite{wen2024foundationpose}, sampling for pose hypothesis generation was only performed in $SO(3)$ without considering size variations. Instead, we additionally sample different sizes, together with $SO(3)$ sampling. In particular, the size samples are drawn in the range of $\Delta s\in [s_0, s_1]$ (we empirically set $s_0=0.6, s_1=1.4$) along each axis. We then refine the sampled pose hypothesis using the refinement module provided by \cite{wen2024foundationpose} and render them to compare with the query image observation. The optimal pose hypothesis is selected based on the comparison score indicated by the pose selection module in \cite{wen2024foundationpose}. 
Once the optimal size is determined, we scale the object shape and switch to the pose refinement stage, including an axis alignment step for more precise accuracy. This updated alignment leverages the joint estimation of size and pose, leading to greater accuracy than traditional IoU-based methods.

After refining the object parameters, we determine the final object pose $T_{{O_M} \rightarrow A}$, which provides an accurate alignment of the reconstructed object shape. With the anchor image $I_A$, query image $I_Q$, and metric-scale object shape ${O}_{M}$, we estimate the relative pose $\mathbf{T}_{A \rightarrow Q}$ by composing two transformations: from the object to the anchor ($\mathbf{T}_{O_M \rightarrow A}$) and from the object to the query ($\mathbf{T}_{O_M \rightarrow Q}$). The relative pose can be expressed as follows:
\begin{equation}
\label{eq:relative_pose}
\mathbf{T}_{A \rightarrow Q} = (\mathbf{T}_{O_M \rightarrow A})^{-1} \cdot \mathbf{T}_{O_M \rightarrow Q}
\end{equation}

For pose selection, we employ a two-level render-and-compare strategy. Initially, a pose ranking network evaluates each hypothesis by comparing its rendered view to the cropped observation, producing an embedding to quantify alignment quality. Then, we apply self-attention to the concatenated embeddings of all hypotheses, incorporating global context to generate final scores for selecting the optimal pose.

\section{Experiments}\label{sec:experiments}

\subsection{Datasets}
We evaluate our method on five diverse real-world datasets: HO3D~\cite{hampali2020honnotate}, YCBInEOTA~\cite{wen2020se}, Toyota-Light~\cite{hodan2018bop},  REAL275~\cite{wang2019normalized}, and LM-O~\cite{brachmann2014learning}. Each dataset presents unique challenges in object pose estimation under various interaction scenarios and environmental conditions.

\begin{table*}[t]
\caption{Model-free pose estimation results measured by AUC of ADD, and ADD-S, AR on HO3D dataset.}
\vspace{-0.1in}
\centering
\def\mywidth{0.85\linewidth} 
\definecolor{grey}{RGB}{230,230,230}
\resizebox{\mywidth}{!}{
\begin{tabular}{l c c c c c c c c c c c c c c c}
\thickline
                               & \multicolumn{3}{c}{Oryon~\cite{corsetti2024open}}  &  & \multicolumn{3}{c}{LoFTR~\cite{sun2021loftr}}          &  & \multicolumn{3}{c}{Gedi~\cite{poiesi2022learning}}  &  & \multicolumn{3}{c}{\cellcolor{grey} Ours}  \\
\cline{2-4} \cline{6-8} \cline{10-12} \cline{14-16}
Modality                    & \multicolumn{3}{c}{RGB-D \& Language}                                         &  & \multicolumn{3}{c}{RGB-D}            &  & \multicolumn{3}{c}{Depth}                                          & & \multicolumn{3}{c}{\cellcolor{grey} RGB-D}  \\
Metrics                        & ADD-S  & ADD   & AR  &  & ADD-S  & ADD  & AR  &  & ADD-S  & ADD  & AR  &  & \cellcolor{grey} ADD-S  & \cellcolor{grey} ADD  & \cellcolor{grey} AR   \\
\hline
AP10 & 23.8 & 0.0 & 0.4 &  & 22.5 & 0.0 & 1.2 &  & 94.4 & 1.9 & 3.5 &  & \cellcolor{grey} 100.0 & \cellcolor{grey} 16.2 & \cellcolor{grey} 22.2 \\
AP11 & 25.6 & 0.0 & 1.3 &  & 59.4 & 15.6 & 14.8 &  & 100.0 & 55.0 & 32.3 &  & \cellcolor{grey} 100.0 & \cellcolor{grey} 73.8 & \cellcolor{grey} 59.0 \\
AP12 & 21.2 & 0.0 & 1.4 &  & 12.5 & 1.2 & 2.1 &  & 99.4 & 30.6 & 20.3 &  & \cellcolor{grey} 100.0 & \cellcolor{grey} 48.8 & \cellcolor{grey} 28.3 \\
AP13 & 26.2 & 0.0 & 0.6 &  & 31.9 & 1.9 & 1.9 &  & 100.0 & 13.1 & 8.8 &  & \cellcolor{grey} 100.0 & \cellcolor{grey} 74.4 & \cellcolor{grey} 45.0 \\
AP14 & 8.1 & 0.0 & 0.0 &  & 25.0 & 0.0 & 0.0 &  & 76.2 & 0.0 & 0.6 &  & \cellcolor{grey} 100.0 & \cellcolor{grey} 35.6 & \cellcolor{grey} 29.7 \\
SM1 & 24.7 & 0.0 & 1.1 &  & 52.8 & 3.4 & 1.9 &  & 82.0 & 0.0 & 1.6 &  & \cellcolor{grey} 86.5 & \cellcolor{grey} 34.8 & \cellcolor{grey} 27.8 \\
SB11 & 46.1 & 0.0 & 2.4 &  & 75.4 & 8.4 & 15.6 &  & 96.4 & 13.8 & 12.1 &  & \cellcolor{grey} 100.0 & \cellcolor{grey} 86.8 & \cellcolor{grey} 68.9 \\
SB13 & 29.9 & 0.0 & 4.2 &  & 33.5 & 0.0 & 1.9 &  & 98.2 & 11.4 & 9.4 &  & \cellcolor{grey} 99.4 & \cellcolor{grey} 64.1 & \cellcolor{grey} 54.6 \\
MPM10 & 8.3 & 0.0 & 0.2 &  & 13.4 & 0.0 & 0.3 &  & 29.9 & 0.0 & 3.1 &  & \cellcolor{grey} 98.7 & \cellcolor{grey} 26.8 & \cellcolor{grey} 31.3 \\
MPM11 & 33.8 & 0.0 & 0.1 &  & 26.1 & 0.0 & 0.0 &  & 35.0 & 0.0 & 0.6 &  & \cellcolor{grey} 100.0 & \cellcolor{grey} 3.2 & \cellcolor{grey} 32.4 \\
MPM12 & 17.2 & 0.0 & 0.1 &  & 5.1 & 0.0 & 0.0 &  & 42.0 & 0.0 & 0.4 &  & \cellcolor{grey} 100.0 & \cellcolor{grey} 1.3 & \cellcolor{grey} 23.5 \\
MPM13 & 24.2 & 0.0 & 0.4 &  & 10.2 & 0.0 & 0.3 &  & 45.2 & 0.0 & 1.3 &  & \cellcolor{grey} 100.0 & \cellcolor{grey} 15.9 & \cellcolor{grey} 30.9 \\
MPM14 & 9.6 & 0.0 & 1.4 &  & 15.9 & 0.0 & 2.0 &  & 35.7 & 0.0 & 1.5 &  & \cellcolor{grey} 98.7 & \cellcolor{grey} 43.9 & \cellcolor{grey} 44.0 \\
\hline
MEAN & 23.0 & 0.0 & 1.0 &  & 29.5 & 2.3 & 3.2 &  & 71.9 & 9.7 & 7.4 &  & \cellcolor{grey} \textbf{98.7} & \cellcolor{grey} \textbf{40.4} & \cellcolor{grey} \textbf{38.3} \\
\thickline
\end{tabular}%
}
\label{tab:ho3d_details_pose}
\vspace{-0.15in}
\end{table*}

\noindent\textbf{HO3D}~\cite{hampali2020honnotate} captures close-range RGB-D images of human-hand object interactions. Using the latest HO3D-V3 version, we evaluate our method on 4 objects across 13 video sequences, totaling 2K images. The dataset presents large variations in viewpoints and significant hand-based occlusions. We sample pair images from the DexYCB~\cite{chao2021dexycb} dataset.

\noindent\textbf{YCBInEOTA}~\cite{wen2020se} features mid-range RGBD images of dual-arm robot manipulations. The dataset includes three types of interactions: single-arm pick-and-place, within-hand manipulation, and pick-and-place with handoff between arms. We evaluate our method on 5 objects across 9 videos, totaling 749 frames, with pair images sampled from DexYCB~\cite{chao2021dexycb}. The dataset presents diverse viewpoints and robot arm occlusions.

\noindent\textbf{Toyota-Light}~\cite{hodan2018bop} focuses on single-object pose estimation under challenging lighting conditions. It features significant lighting variations between scenes, which present crucial challenges. Following Oryon\cite{corsetti2024open}, we evaluate our method on 2K image pairs.

\noindent\textbf{REAL275}~\cite{wang2019normalized} consists of RGBD images captured across different scenes, featuring six object categories with three instances per category. The dataset exhibits limited viewpoint variations and includes scenarios with minor occlusions. Following Oryon\cite{corsetti2024open}, we evaluate our method on 2K image pairs selected from the original real-world test set.

\noindent\textbf{LM-O}~\cite{brachmann2014learning} consists of RGBD images captured in cluttered scenes, featuring 12 distinct textureless objects. The dataset has viewpoint variations and realistic occlusion scenarios. We follow GigaPose\cite{nguyen2024gigapose} of input image, segmentation, and image-to-3d methods for comparison.

\subsection{Metrics}
Our primary evaluation metrics are based on the Average Distance of Model Points (ADD)~\cite{xiang2017posecnn, wang2019densefusion}, which computes the mean distance between corresponding 3D model points under-predicted and ground truth poses. For asymmetric objects, we use ADD directly, while for symmetric objects, we employ ADD-S, which calculates the average distance to the closest model point. We report the Area Under the Curve (AUC) and the recall rate at a threshold of 0.1 times the object diameter~\cite{he2022fs6d, he2022oneposepp}.
Additionally, we evaluate using the metrics established in the BOP challenge~\cite{hodan2024bop}, including the Average Recall (AR) of Visual Surface Discrepancy (VSD), Maximum Symmetry-aware Surface Distance (MSSD), and Maximum Symmetry-aware Projection Distance (MSPD).  These metrics provide complementary perspectives on pose accuracy by evaluating recalls over multiple thresholds.

\subsection{Comparison with State-of-the-art}\label{subsec:sota_comparision}
The evaluation is conducted on 5 challenging datasets: HO3D, YCBInEOTA, Toyota-Light, REAL275, and LM-O, each presenting unique challenges for pose estimation. For evaluation, we align the relative pose $T_{A \rightarrow Q}$ with the object pose $T_{O \rightarrow Q}$ by multiplying it with $T_{O \rightarrow A}$, following the approach used in Oryon~\cite{corsetti2024open}. We use ground-truth segmentation masks to evaluate the HO3D, YCBInEOTA, Toyota-Light, and REAL275 datasets. 

We compare our approach against several recent state-of-the-art methods, including Oryon~\cite{corsetti2024open}, a multi-modal object pose estimation method, and single-image matching baselines such as LoFTR~\cite{sun2021loftr}, Gedi~\cite{poiesi2022learning}, and ObjectMatch~\cite{gumeli2023objectmatch}. For LoFTR, we use matched point clouds combined with pose optimization~\cite{umeyama1991least} to estimate accurate object poses. ObjectMatch leverages SuperGlue~\cite{sarlin20superglue} for match estimation, as outlined in Oryon~\cite{corsetti2024open}. These baselines cover a range of approaches to pose estimation, from traditional feature matching to recent learning-based methods.

\begin{table*}[ht]
\caption{Model-free pose estimation results measured by AUC of ADD, ADD-S, and AR on YCBINEOAT dataset.}
\vspace{-0.1in}
\centering
\def\mywidth{0.9\linewidth} 
\definecolor{grey}{RGB}{230,230,230}
\resizebox{\mywidth}{!}{
\begin{tabular}{l c c c c c c c c c c c c c c c}
\thickline
                               & \multicolumn{3}{c}{Oryon~\cite{corsetti2024open}}  &  & \multicolumn{3}{c}{LoFTR~\cite{sun2021loftr}}          &  & \multicolumn{3}{c}{Gedi~\cite{poiesi2022learning}}  &  & \multicolumn{3}{c}{\cellcolor{grey} Ours}  \\
\cline{2-4} \cline{6-8} \cline{10-12} \cline{14-16}
Modality                    & \multicolumn{3}{c}{RGB-D \& Language}                                         &  & \multicolumn{3}{c}{RGB-D}            &  & \multicolumn{3}{c}{Depth}                                          & & \multicolumn{3}{c}{\cellcolor{grey} RGB-D}  \\
Metrics                        & ADD-S  & ADD  & AR  &  & ADD-S  & ADD  & AR  &  & ADD-S  & ADD  & AR  &  & \cellcolor{grey} ADD-S  & \cellcolor{grey} ADD  & \cellcolor{grey} AR   \\
\hline
sugar$\_$box1 & 44.0 & 0.0 & 1.1 &  & 47.3 & 0.0 & 0.1 &  &95.6 &0.0 &1.7 &  & \cellcolor{grey} 96.7 & \cellcolor{grey} 14.3 & \cellcolor{grey} 11.3 \\
sugar$\_$box$\_$yalehand0 & 34.7 & 3.0 & 5.2 &  & 41.6 & 0.0 & 0.1 &  &82.2 &6.9 &21.5 &  & \cellcolor{grey} 89.1 & \cellcolor{grey} 75.2 & \cellcolor{grey} 44.4 \\
mustard0 & 48.6 & 0.0 & 3.5 &  & 47.3 & 20.3 & 15.2 &  &100 &0.0 &19.1 &  & \cellcolor{grey} 100 & \cellcolor{grey} 23 & \cellcolor{grey} 32.4 \\
mustard$\_$easy$\_$00$\_$02 & 36.2 & 0.0 & 0.3 &  & 23.2 & 0.0 & 1.9 &  &78.3 &0.0 &20.2 &  & \cellcolor{grey} 78.3 & \cellcolor{grey} 53.6 & \cellcolor{grey} 39.2 \\
bleach0 & 10.4 & 0.0 & 1.5 &  & 55.2 & 0.0 & 0.9 &  &74.6 &0.0 &7.7 &  & \cellcolor{grey} 98.5 & \cellcolor{grey} 68.7 & \cellcolor{grey} 56 \\
bleach$\_$hard$\_$00$\_$03$\_$chaitanya & 24.4 & 6.7 & 6.1 &  & 60 & 15.6 & 18.7 &  &66.7 &62.2 &35.5 &  & \cellcolor{grey} 73.3 & \cellcolor{grey} 51.1 & \cellcolor{grey} 37.7 \\
tomato$\_$soup$\_$can$\_$yalehand0 & 32.8 & 0.0 & 4.5 &  & 10.7 & 0.0 & 6.8 &  &60.3 &0.0 &7.8 &  & \cellcolor{grey} 70.2 & \cellcolor{grey} 0 & \cellcolor{grey} 14.1 \\
cracker$\_$box$\_$reorient & 13.2 & 0.0 & 0 &  & 26.3 & 0.0 & 0 &  &97.4 &0.0 &1.8 &  & \cellcolor{grey} 100 & \cellcolor{grey} 60.5 & \cellcolor{grey} 44.2 \\
cracker$\_$box$\_$yalehand0 & 15.0 & 0.0 & 1.2 &  & 22.6 & 0.0 & 0.2 &  &89.5 &0.0 &10.4 &  & \cellcolor{grey} 97.7 & \cellcolor{grey} 63.9 & \cellcolor{grey} 58.2 \\
\hline
MEAN & 28.8 & 1.1 & 2.6 &  & 37.1 & 4 & 4.9 &  & 82.7 & 7.7 & 14.0 &  & \cellcolor{grey} \textbf{89.3} & \cellcolor{grey} \textbf{45.6} & \cellcolor{grey} \textbf{37.5} \\
\thickline
\end{tabular}%
}
\vspace{-0.15in}
\label{tab:ycbineoat_details_pose}
\end{table*}

\Tref{tab:ho3d_details_pose} summarized the results on HO3D dataset. The experimental results on the HO3D dataset demonstrate that our proposed method significantly outperforms state-of-the-art approaches across all evaluation metrics. Specifically, our method achieves mean scores of 98.7\%, 40.4\%, and 38.3\% for ADD-S, ADD, and AR metrics, respectively, substantially surpassing previous methods, including Oryon (4.1\%, 0\%, 0.2\%), LoFTR (29.5\%, 2.3\%, 3.2\%), and Gedi (71.9\%, 9.7\%, 7.4\%).  These results are particularly notable given our method's consistent performance across object instances, even in challenging scenarios involving human-hand interactions and occlusions. The exceptional improvement in ADD-S metrics, approaching 100\% in most cases, validates our method's robustness and accurate pose estimation under dynamic conditions.

For the YCBInEOAT dataset (Table~\ref{tab:ycbineoat_details_pose}), our approach achieves superior performance with mean scores of 89.3, 45.6, and 37.5 for ADD-S, ADD, and AR, significantly surpassing the corresponding scores of Gedi, which are 82.7, 7.7, and 14.0. The substantial improvement in ADD metrics particularly highlights our method's capability in precise pose estimation. These results validate our method's effectiveness in challenging scenarios with occlusions and non-overlapping viewpoints, which are critical for real-world robotic applications.

\begin{table}[tb]
\caption{Model-free pose estimation results measured by AUC of ADD(-S), AR, MSSD, MSPD, and VSD on the Toyota-Light (TOYL) dataset.}
\vspace{-0.1in}
\centering
\resizebox{0.9\linewidth}{!}{  
\begin{tabular}{lccccc}
\thickline
Method & ADD(-S)  & AR  & MSSD  & MSPD  & VSD  \\ 
\hline
SIFT~\cite{lowe1999object}            & 14.1         & 30.3        & 39.6          & 44.1          & 7.3         \\ 
Obj. Mat.~\cite{gumeli2023objectmatch}    & 5.4          & 9.8        & 13.0          & 14.0          & 2.4         \\ 
Oryon~\cite{corsetti2024open}           & 22.9          & 34.1        & 42.9          & 45.5          & 13.9         \\ 
\rowcolor[gray]{0.9} % Sets the row background to light grey
Ours            & \textbf{32.2}          & \textbf{43.3}        & \textbf{55.8}          & \textbf{58.4}          & \textbf{15.8}         \\ 
\thickline
\end{tabular}
}
\vspace{-0.05in}
\label{tab:toyl_details_pose}
\end{table}

On the Toyota-Light dataset (Table~\ref{tab:toyl_details_pose}), our method demonstrates substantial improvements across all metrics. We achieve 32.2\% in ADD(-S), 43.3\% in AR, 55.8\% in MSSD, and 58.4\% in MSPD, consistently outperforming Oryon by significant margins (9.3\%, 9.2\%, 12.9\%, and 12.9\% respectively). Our approach shows particular robustness under varying lighting conditions, which often pose significant challenges for existing methods. The consistent performance across all metrics demonstrates the stability of our approach in handling different lighting scenarios.

\begin{table}[tb]
\caption{Model-free pose estimation results measured by AUC of ADD(-S), AR, MSSD, MSPD, and VSD on the REAL275 dataset.}
\vspace{-0.1in}
\centering
\resizebox{0.9\linewidth}{!}{  
\begin{tabular}{lccccc}
\thickline
Method & ADD(-S)  & AR  & MSSD  & MSPD  & VSD  \\ \hline
SIFT~\cite{lowe1999object}            & 16.4          & 34.1        & 37.9          & 48.0          & 16.5         \\ 
Obj. Mat.~\cite{gumeli2023objectmatch}    & 13.4          & 26.0        & 31.7          & 30.8          & 15.5         \\ 
Oryon~\cite{corsetti2024open}           & 34.9          & 46.5        & 50.9          & 56.7          & \textbf{32.1}         \\ 
\rowcolor[gray]{0.9}
Ours            & \textbf{53.5}          & \textbf{51.0}        & \textbf{56.5}          & \textbf{65.3}          & 31.1         \\ 
\thickline
\end{tabular}
}
\vspace{-0.15in}
\label{tab:nocs_details_pose}
\end{table}

\begin{table}[tb]
\caption{Model-free pose estimation results measured by AUC of AR, MSSD, MSPD, and VSD on the Linemod Occlusion (LM-O) dataset.}
    \vspace{-0.10in}
    \centering
    \resizebox{1.0\linewidth}{!}{
    \begin{tabular}{l c c c c c c c}
    \thickline
    \multirow{2}{*}{Method} & \multirow{2}{*}{Segmentation} & \multirow{2}{*}{Image-to-3D} & \multicolumn{4}{c}{Metrics} \\
    \cline{4-7}
    & & & AR  & MSPD  & MSSD  & VSD   \\ \hline
    GigaPose~\cite{nguyen2024gigapose} & CNOS~\cite{nguyen2023cnos} & Wonder3D~\cite{long2024wonder3d} & 17.5 & 35.8 & 9.0 & 7.6 \\
    \rowcolor[gray]{0.9} Ours & CNOS~\cite{nguyen2023cnos} & Wonder3D~\cite{long2024wonder3d} & \textbf{28.6} & \textbf{36.1} & \textbf{32.0} & 17.6 \\
    \rowcolor[gray]{0.9} Ours & CNOS~\cite{nguyen2023cnos} & InstantMesh~\cite{xu2024instantmesh} & 25.2 & 29.5 & 27.4 & \textbf{18.7} \\
    \thickline
    \vspace{-0.2in}
    \end{tabular}
}
    \label{tab:linemod_details_pose}
    \end{table}

\begin{figure*}[ht]
\begin{center}
\includegraphics[width=1.0\linewidth]{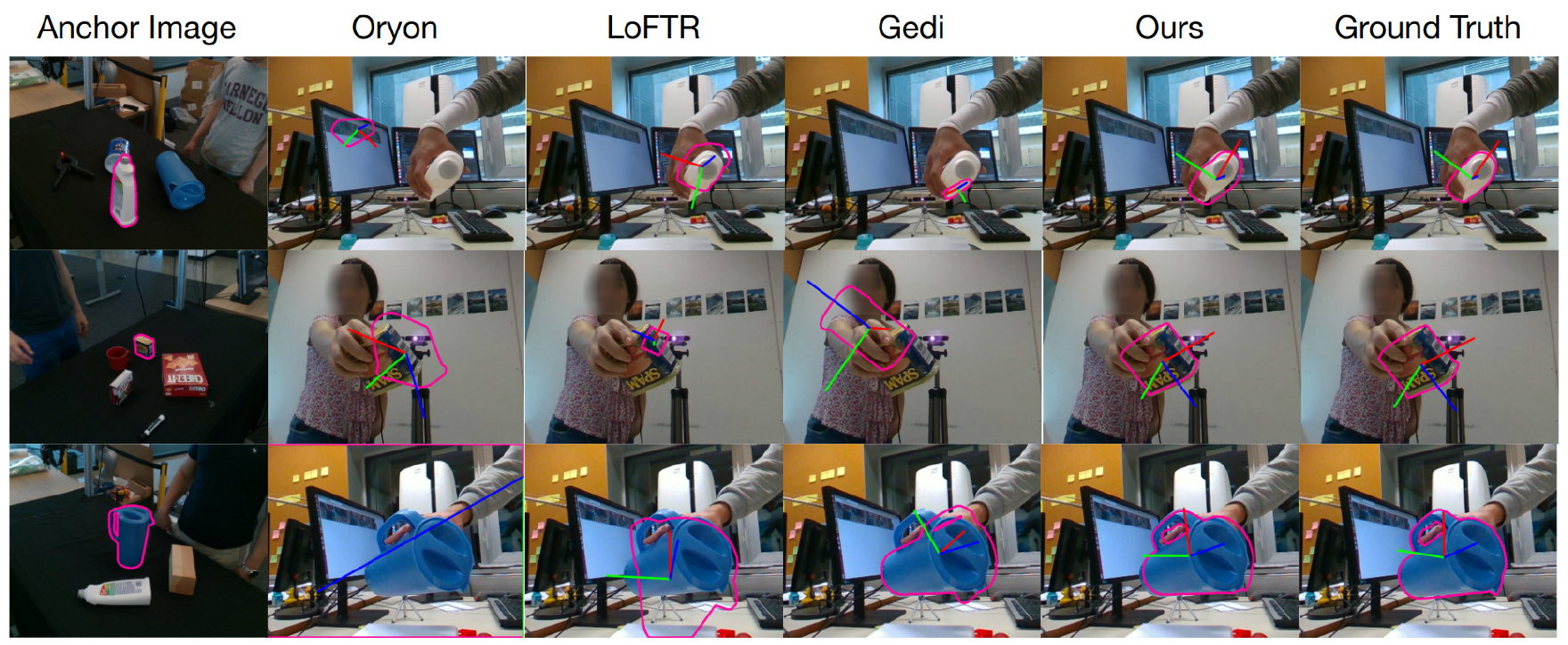}
\vspace{-0.15in}
\caption{Qualitative comparison of state-of-the-art methods on the HO3D Dataset. In this challenging scenario, the left anchor image shows only partially visible objects, while the query images are not visible due to occlusion or different viewing angles. This represents the most challenging case for matching. Gedi, being a depth-based method, shows ambiguity when dealing with RGB-based non-symmetric objects.}
\label{fig:ho3d_qual}
\end{center}
\vspace{-0.25in}
\end{figure*}

On the REAL275 dataset (Table~\ref{tab:nocs_details_pose}), our method demonstrates remarkable performance, achieving 53.5\% in ADD(-S), 51.0\% in AR, 56.5\% in MSSD, and 65.3\% in MSPD. These results significantly surpass previous methods, particularly showing substantial improvements over Oryon in most metrics (improvements of 18.6\%, 4.5\%, 27.1\%, and 33.8\%, respectively). While maintaining competitive performance in VSD (31.1\% versus Oryon's 32.1\%), our method excels in all other metrics, demonstrating superior generalization across various object categories and poses. This comprehensive evaluation validates the robustness and versatility of our approach in handling diverse real-world scenarios.

Finally, on the Linemod Occlusion (LM-O) dataset (Table~\ref{tab:linemod_details_pose}), we compare the estimation of the pose with a 3D model predicted from a single image. We compare the same input images, segmentation~\cite{nguyen2023cnos}, and image-to-3D~\cite{long2024wonder3d} followed by GigaPose~\cite{nguyen2024gigapose} for fair comparison. Our method outperforms GigaPose under the same settings. Note that Gigapose is an RGB-based method but requires the initial pose to determine object size, while our method estimates object size from a single RGB-D image.

These extensive experiments across multiple datasets demonstrate that our method consistently outperforms existing approaches, often by significant margins. The strong performance across different metrics and various challenging scenarios validates the effectiveness and reliability of our approach for real-world applications.

\subsection{Qualitative Results}

\Fref{fig:ho3d_qual} shows qualitative results of our method compared to Oryon~\cite{corsetti2024open}, LoFTR~\cite{sun2021loftr}, and Gedi~\cite{poiesi2022learning} on the HO3D dataset. Given the anchor image (left), we overlay the object contours in \pink{pink}, along with the rendered object and rotation axes, to visualize the pose estimation results. The query images are selected to have no overlapping parts with the anchor images, creating a challenging scenario for pose estimation. While Oryon, LoFTR, and Gedi struggle under these conditions due to their reliance on partial anchor information, our method effectively reconstructs complete object shapes, allowing for robust pose estimation even when parts of the object are occluded or not visible in the anchor view.

The first example features a white cleanser object where only a portion is visible in the anchor image, with a significantly different viewpoint in the query image. Our method accurately estimates the pose, while others fail to handle the limited visibility of key features. In the second example with a SPAM can, the anchor image shows only the back view, while the query image captures the logo side partially occluded by a hand. Despite these challenging conditions, our method successfully estimates the correct pose, while competing methods fail to align due to their reliance on limited anchor information. The final example shows a blue pitcher where Gedi roughly estimates the center position but fails to capture the correct orientation. In contrast, our method successfully aligns rotational and translational components, closely matching the ground truth pose.

\begin{figure*}[t]
\begin{center}
\includegraphics[width=1.0\linewidth]{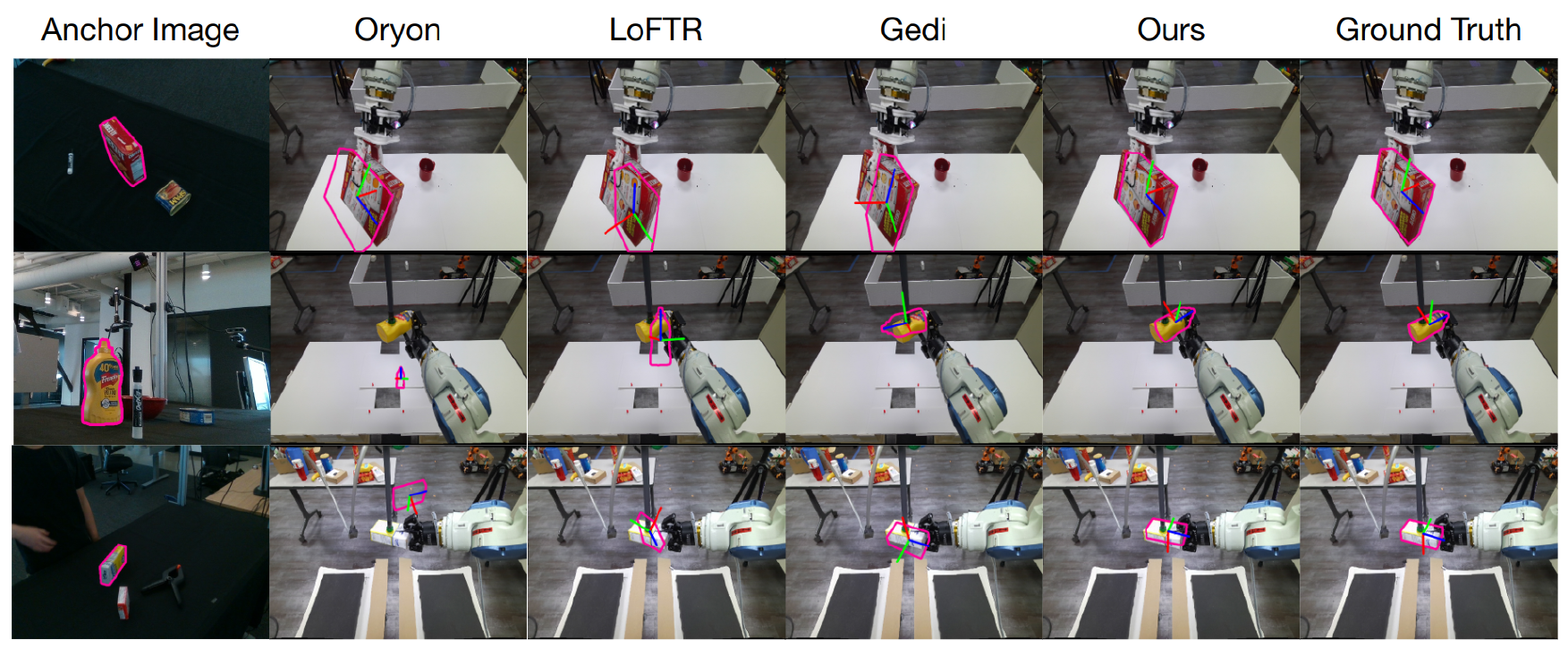}
\vspace{-0.2in}
\caption{Qualitative comparison of state-of-the-art methods on the YCBInEOAT Dataset. In this challenging scenario, the left anchor image shows only partially visible objects, while the query images are not visible due to occlusion or different viewing angles. This represents the most challenging case for matching. Gedi, being a depth-based method, shows ambiguity when dealing with RGB-based non-symmetric objects.}
\label{fig:ycbineoat_qual}
\end{center}
\vspace{-0.2in}
\end{figure*} 

\begin{figure}[ht]
\begin{center}
\includegraphics[width=1.0\linewidth]{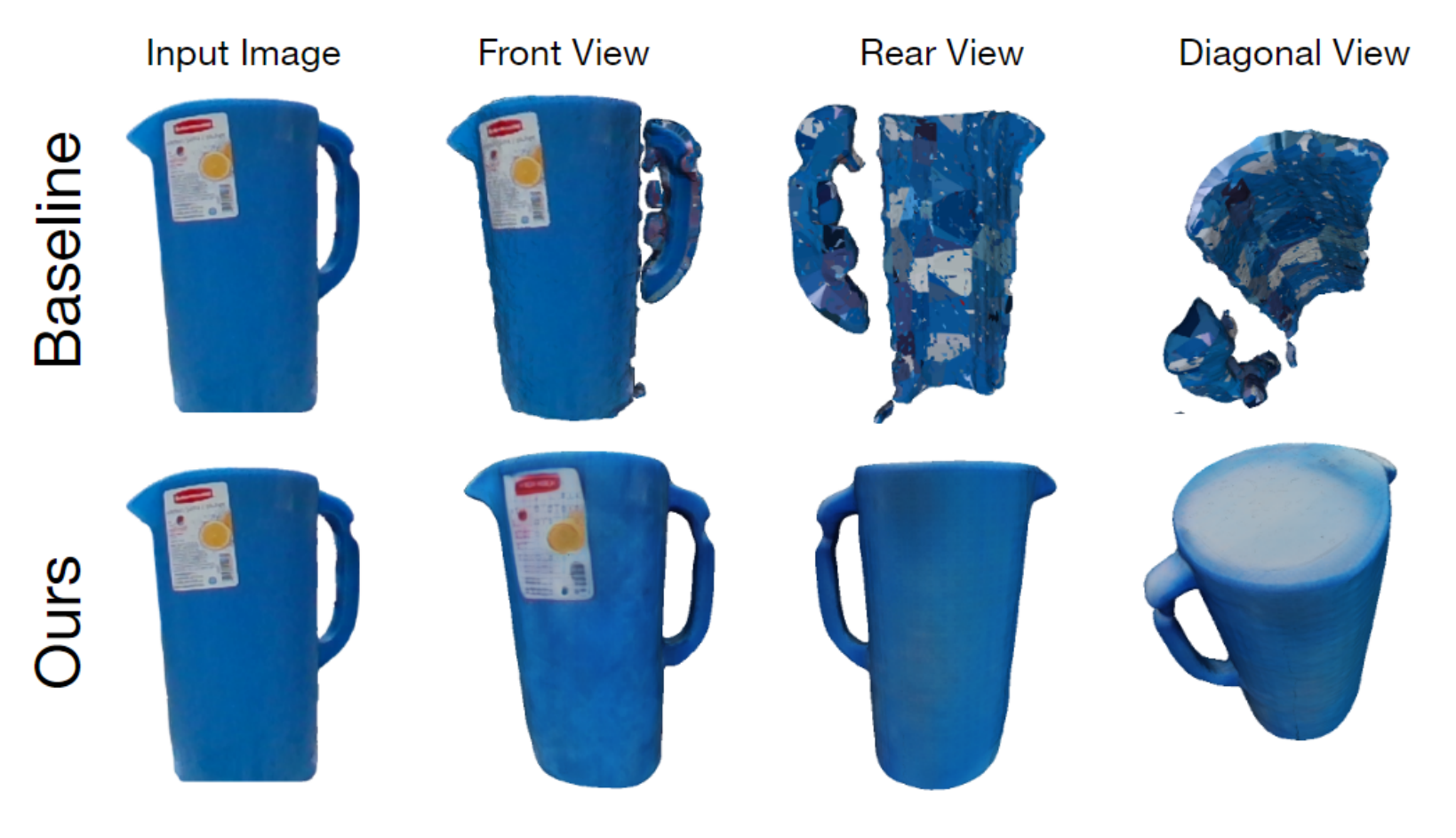}
\vspace{-0.1in}
\caption{Comparison of shape quality between baseline method and ours.
}
\label{fig:ablation_study}
\end{center}
\vspace{-0.2in}
\end{figure} 

\Fref{fig:ycbineoat_qual} demonstrates our pose estimation results in robotic manipulation scenarios on the YCBInEOAT dataset. We compare our method with Oryon, LoFTR, and Gedi across various examples of robot-object interaction. In the first example with a red cracker box, our method accurately aligns both position and orientation, while other methods struggle with pose estimation. Notably, the anchor image only shows the front cheese logo, yet our method successfully handles the back view in the query image. The second example involves a yellow mustard bottle manipulated by two dual robot arms. Our method maintains accurate pose estimation throughout the interaction, proving robust to occlusions from the gripper. In the final example with a yellow sugar box, our method achieves precise pose estimation despite minimal visible information in the anchor image, while competing methods fail to handle the challenging viewpoint variations.
These qualitative results highlight our method handles challenging scenarios involving significant occlusions and viewpoint changes, consistently outperforming existing approaches in pose estimation accuracy.

\subsection{Ablation Studies}\label{ablation_studies}
In this section, we conduct ablation studies to evaluate our object shape and alignment estimation.

\noindent\textbf{Object Shape.} We evaluate different object reconstruction approaches for pose estimation, including a comparison with a partial view-based baseline. For this baseline, given a camera-to-object pose $T_{C\rightarrow O}$ and a reference RGB-D image $I_r$, we train an object-centric NeRF~\cite{wen2023bundlesdf} model and use marching cubes~\cite{we1987marching} to extract a textured mesh. This baseline relies on ground-truth poses to train NeRF with aligned object coordinates, ensuring a fair comparison with the same pose estimation setup. It follows the single image version of FoundationPose~\cite{wen2024foundationpose}.
As shown in \Fref{fig:ablation_study}, our reconstructed shapes differ significantly from the baseline, which exhibits missing regions in rear and diagonal views, leading to ambiguity in rendering and pose estimation. \Tref{tab:ho3d_ablation_studies} presents the pose estimation results and mesh quality evaluation using Chamfer Distance (CD). The CD between our reconstructions and ground-truth meshes indicates that lower values correspond to better accuracy. Additionally, we found that improper axis alignment leads to distortions in the X, Y, and Z ratios, highlighting the importance of precise axis alignment.

\begin{table}[tb]
    \caption{Ablation Studies of Size Estimation on the HO3D dataset. }\vspace{-7pt}
    \centering
    \resizebox{1.0\linewidth}{!}{
    \begin{tabular}{l c c c c c c c c c}
\thickline
     \multirow{2}{*}{Method}  & & \multicolumn{3}{c}{Object Alignment} & & \multicolumn{4}{c}{Metrics}  \\
     & &Coarse Size &  Refinement & Axis Align  & & ADD-S ($\uparrow$) & ADD ($\uparrow$) & AR ($\uparrow$) & CD ($\downarrow$) \\
    \cline{1-1} \cline{3-5} \cline{7-10}
    Baseline & &\redxmark &\redxmark &\redxmark & & 28.6 & 0.00 & 0.20 & 1.02\\
        (1) & &\redxmark & \redxmark & \redxmark & & 0.0 & 0.0 & 0.0 & 1.47 \\
        (2) & &\redxmark & \greencheckmark & \greencheckmark & & 98.0 & 25.5 & 26.8 & 0.53 \\
        (3) & &\greencheckmark & \redxmark & \greencheckmark & & 83.7 & 26.6 & 22.5 & 0.92 \\
        (4) & &\greencheckmark & \greencheckmark & \redxmark & & 92.3 & 23.6 & 24.9 & 0.66 \\
        \rowcolor[gray]{0.9} Ours & &\greencheckmark & \greencheckmark & \greencheckmark & & \textbf{98.7} & \textbf{40.4} & \textbf{38.3} & \textbf{0.49} \\
        \thickline
    \end{tabular}}
    \label{tab:ho3d_ablation_studies}
    \vspace{-0.1in}
\end{table}

\noindent\textbf{Object Alignment.} We evaluate our simple but effective object alignment module: coarse size estimation, size and pose refinement and axis alignment. The results of our ablation study on these components are presented in \Tref{tab:ho3d_ablation_studies}. First, as shown in \Tref{tab:ho3d_ablation_studies}-(1), using raw object shape estimation without any alignment steps leads to failure in accurately estimating the object’s rotation and translation, with all metrics showing subpar performance. This underscores the importance of size alignment as a foundational step in pose estimation. In \Tref{tab:ho3d_ablation_studies}-(2) and in our full method, incorporating coarse size estimation substantially improves the ADD metric. This shows that even a basic size estimation allows the model to approximate the object’s pose better. Next, \Tref{tab:ho3d_ablation_studies}-(3) demonstrates that incorporating axis alignment further enhances performance, particularly on the ADD-S and AR metrics. This process not only improves object shape estimation but also yields significant gains in pose accuracy by aligning the object’s axes to avoid distortion in its proportions along the x, y, and z directions. Finally, our full method, incorporating coarse size estimation, refinement, and axis alignment, achieves the best results across all metrics. Specifically, as shown in \Tref{tab:ho3d_ablation_studies}, our method reaches an ADD of 40.4 and an AR of 38.3, outperforming the other configurations. These results validate the effectiveness of our alignment approach in enhancing object pose estimation.

\section{Conclusion}
We introduce Any6D, a novel framework for model-free object pose estimation that reduces dependence on CAD models and multi-view images, particularly in challenging object manipulation scenarios. Our method proposes an efficient object alignment method for precise pose and size estimation. Extensive experiments demonstrate that Any6D significantly outperforms state-of-the-art methods for occlusions and varying viewpoints. 
While our method shows promising results on pose and size estimation through image-to-3D alignment, it has limitations when the initial 3D shape is inaccurate, as our approach does not incorporate shape updating. A future direction would be to refine shape to enhance robustness and applicability in scenarios with inaccurate initial shapes.

\section*{Acknowledgment}
This work was partly supported by the National Research Foundation of Korea(NRF) grant funded by the Korea government(MSIT) (NRF2022R1A2B5B03002636), the Institute of Information and Communications Technology Planning and Evaluation (IITP) grant funded by the Korea Government (MSIT) (Artificial Intelligence Innovation Hub) under Grant 2021-0-02068, and KAIST Cross-Generation Collaborative Lab Project.
% \newpage

{
    \small
    \bibliographystyle{ieeenat_fullname}
    \bibliography{main}
}


\end{document}